\pdfoutput=1
\documentclass[11pt]{article}
\usepackage[letterpaper,margin=1in]{geometry}
\usepackage[T1]{fontenc}
\usepackage[utf8]{inputenc}
\usepackage{mathptmx}
\usepackage{amsmath,amssymb}
\usepackage{graphicx}
\usepackage{booktabs,longtable,array,calc}
\usepackage[font=small,labelfont=bf]{caption}
\usepackage{xcolor}
\usepackage{microtype}
\usepackage{newunicodechar}
\newunicodechar{−}{\ensuremath{-}}
\newunicodechar{≥}{\ensuremath{\geq}}
\newunicodechar{≤}{\ensuremath{\leq}}
\newunicodechar{→}{\ensuremath{\rightarrow}}
\newunicodechar{∂}{\ensuremath{\partial}}
\newunicodechar{±}{\ensuremath{\pm}}
\newunicodechar{×}{\ensuremath{\times}}
\newunicodechar{⁷}{\textsuperscript{7}}
\newunicodechar{⁴}{\textsuperscript{4}}
\newunicodechar{¹}{\textsuperscript{1}}
\usepackage[colorlinks=true,linkcolor=blue!50!black,citecolor=blue!50!black,urlcolor=blue!50!black]{hyperref}
\providecommand{\tightlist}{\setlength{\itemsep}{0pt}\setlength{\parskip}{0pt}}
\title{Imitation Policies Learn a Proprioceptive Anchor:\\ A Predictive Law for Perception Noise and Its Closed-Loop Cost}
\author{Christopher Malili\thanks{Code: \href{https://github.com/cmalili/proprioceptive-anchor}{https://github.com/cmalili/proprioceptive-anchor}}\\ \href{mailto:cmalili@asu.edu}{cmalili@asu.edu}}
\date{}
\begin{document}
\maketitle
\begin{abstract}
Imitation policies for robot manipulation are often trained on noisy object positions, and they learn to discount the noisy reading. We show how much they discount it, and why. Because demonstrators steer toward the object, the robot's own state carries information about where the object is. A trained policy combines this proprioceptive anchor with the reading and weights the reading by \(\alpha = V/(V + s^2)\), where \(s^2\) is the reading noise and \(V\) the anchor's error variance. \(V\) belongs to the learner and is far smaller than a linear model of the demonstrations implies, which is why classical errors-in-variables (EIV) regression overestimates \(\alpha\). Measured once, \(V\) predicted ten held-out sensitivities across two robomimic tasks, two learners and four new noise levels, with mean absolute error 0.025 against 0.37 for classical EIV, and a parameter-free bias test rules out the simpler explanation that the policies just overestimate the noise. The discount does not carry over to deployment, where the policy produces its own state: closed-loop sensitivity to a persistent reading offset exceeded \(\alpha\) in all 15 policies that passed the registered gates, although the stricter registered prediction, that all would follow the offset fully, held for 8 of them. Post hoc, lower \(\alpha\) went with lower success, steeply for diffusion, and a registered prospective test confirmed this for diffusion on Can at two new noise levels: the measured \(\alpha\) predicted success before any rollout, within 0.03 (95\% intervals \(\pm 0.08\)), although training noise predicts it equally well. Every test was registered before its data, including the classical benchmark that failed, and all other analyses are labelled post hoc; all results are in simulation with synthetic Gaussian noise.
\end{abstract}
\hypertarget{introduction}{%
\section{Introduction}\label{introduction}}
\begin{figure}[t]
\centering
\includegraphics[width=\linewidth]{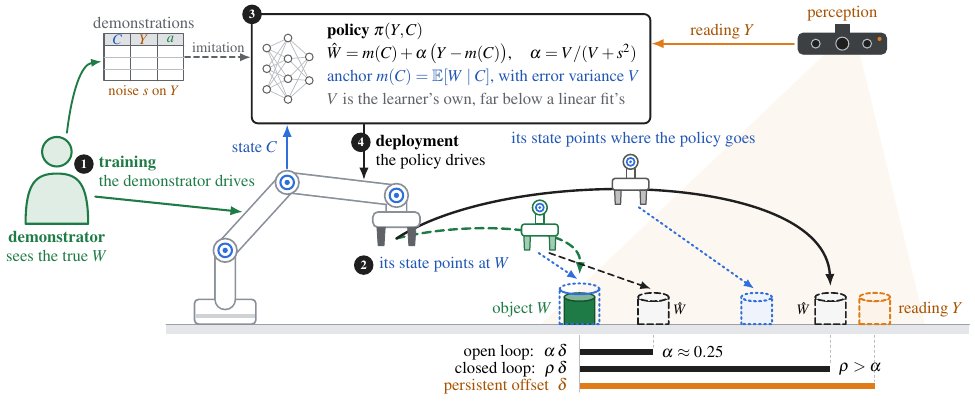}
\caption{Who drives the robot decides what its state says about the object. (1)~In training, a human demonstrator who sees the true position $W$ drives the robot, and each episode is recorded with the robot state $C$, a reading $Y$ with added noise $s$, and the action $a$. (2)~Because the demonstrator steers toward $W$, a demonstration state points at it. (3)~A policy trained by imitation learns this anchor, $m(C) = \mathbb{E}[W \mid C]$ with error variance $V$, and moves its estimate $\hat W$ only a fraction $\alpha = V/(V + s^2)$ of the way toward the reading. (4)~In deployment the policy drives, so its state points where it is going, and until contact a persistent reading offset $\delta$ is indistinguishable from a moved can. Ghost cans mark beliefs about the can's position: blue, the anchor; orange, the reading; black, the estimate. Open loop, the estimate moves $\alpha\,\delta$ (drawn at the $\alpha = 0.25$ measured for diffusion policies on Can at $s = 12$\,mm); closed loop, the aim moved $\rho\,\delta$, with $\rho > \alpha$ in all 15 policies that passed the registered gates (drawn at $\rho = 0.85$; the deployed state's anchor, at $0.6\,\delta$, is illustrative). $V$ belongs to the learner: on Square, diffusion policies act as if $\sqrt{V} \approx 6.5$\,mm, where a linear fit of $W$ on $C$ gives 30\,mm and classical EIV predicts $\alpha = 0.86$ against the measured 0.23. Measured once, $V$ predicted $\alpha$ at four new noise levels in all 10 registered predictions (Figure~\ref{fig:pred}). Lower $\alpha$ went with lower success (post hoc), and a registered test then confirmed this prospectively for diffusion on Can (Figure~\ref{fig:success}).}
\label{fig:overview}
\end{figure}

A robot that learns by imitation usually gets object positions from perception, and those estimates are noisy. Training on noisy observations is a common remedy: the learner sees the noise during training and learns to discount the reading. How much it discounts the reading matters, because that sets how much perception error reaches the actions. Yet no quantitative account predicts this discount for modern policy classes before they are trained.

The obvious candidate is errors-in-variables (EIV) regression. A least-squares learner whose input is measured with noise attenuates its response to that input by the reliability ratio \(\Lambda = \Sigma(\Sigma + s^2 I)^{-1}\), where \(\Sigma\) is the spread of the true input and \(s^2\) the noise variance (Fuller 1987; Carroll et al.~2006). On two toy tasks with scripted experts, where the learner's only clean input was the task phase, this prediction held up: it matched the learned gain on a held-out task and, after one failed attempt, predicted success at new configurations (Appendix~\ref{app:toy}). We then pre-registered it for diffusion policies (Chi et al.~2023) on the robomimic Square task (Mandlekar et al.~2021), with \(\Sigma\) estimated from the demonstrations.

It failed. The predicted ranking of the noise conditions held, but at 12 mm of noise the policies kept 23\% of their clean sensitivity on the informative axis, where EIV predicted 86\%. Rather than stop there, we ran a sequence of diagnostic tests, each with its decision rule written before its data (Section~\ref{sec:eiv}). They ruled out a mis-estimated spread, local memorization, within-demonstration noise structure (tested on the MLP) and recall of training placements.

What remained is a proprioceptive anchor. Demonstrators steer toward the object, so the robot's own state carries information about where the object is. A learner that combines this anchor with the reading, weighting each by its reliability, responds to the reading with \(\alpha = V/(V + s^2)\), where \(V\) is the anchor\textquotesingle s error variance. This is the EIV formula with the right conditioning set. The catch is that \(V\) belongs to the learner: linear and nearest-neighbour estimators of the Square demonstrations put the anchor's error at about 30 mm, where the diffusion policies act as if it were 6 to 7 mm. So we measure \(V\) once, from one noise-trained policy, and predict every other noise level.

Deployment raises a different question, because in closed loop the robot's state is produced by the policy itself. A persistent reading offset is then indistinguishable from a displaced object, and the anchor cannot protect against it. We find that the open-loop protection does not survive, and that policies with lower \(\alpha\) succeed less often. Figure~\ref{fig:overview} summarizes the mechanism.

Contributions:

\begin{enumerate}
\def\labelenumi{\arabic{enumi}.}
\tightlist
\item
  \textbf{A predictive law for sensitivity to perception noise.} With \(V\) measured at one noise level, \(\alpha = V/(V + s^2)\) predicted ten held-out sensitivities across two tasks, two learner classes and four new noise levels. Its mean absolute error was 0.025, with all ten within their registered tolerances (0.10, and 0.05 for one secondary); classical EIV's was 0.37, with none within tolerance (Section~\ref{sec:law}).
\item
  \textbf{A test that rules out noise inflation.} A learner that acts as if the noise were larger, with no anchor, has a bias at clean readings of \(-(1 - \alpha)\). A learner whose anchor is the conditional mean of the object position given the robot state has a small one. Measured biases are small on both tasks and both learners, \(-0.003\) to \(-0.052\) against \(-0.37\) to \(-0.75\) for noise inflation. Their size agrees with the conditional-mean value on Can and, within a wide interval, for the MLP on Square; for diffusion on Square it is smaller (Section~\ref{sec:bias}).
\item
  \textbf{The closed-loop cost.} Sensitivity to a persistent reading offset exceeded \(\alpha\) in all 15 policies that passed the registered gates, so none kept its open-loop protection. The stricter registered prediction, that every policy follows the offset fully, held for 8 of them, so that verdict is mixed. For the MLP the offset response matched the response to a real displacement of the object (r = 0.94, post hoc). For diffusion policies on Can, a registered prospective test at two new noise levels confirmed that open-loop \(\alpha\) predicts closed-loop success, within 0.03 (95\% intervals \(\pm 0.08\)), although a curve in training noise predicts it as well (Section~\ref{sec:closed}).
\item
  \textbf{A complete pre-registration record.} Every test's rule and predicted numbers were written before its data, and the later registrations also record the hashes of the code that ran them. The record includes the failed benchmark and the tests whose registered verdicts were mixed or negative (Appendix A).
\end{enumerate}

Scope: the robot experiments are in simulation (robosuite), with synthetic Gaussian noise on the object position, low-dimensional observations, 200 human demonstrations per task and two seeds per configuration. Section 6 discusses what this does and does not license.

\hypertarget{related-work}{%
\section{Related work}\label{related-work}}

\textbf{Measurement-error models.} Regression on a noisily measured input attenuates the fitted coefficient by the reliability ratio, a result known as regression dilution (Fuller 1987; Frost and Thompson 2000). With additional error-free covariates, the attenuation depends on the variance of the true input given those covariates (Carroll et al.~2006, ch.~3). Our anchor law is that result with the robot state as the covariate. What is new is that, for deep imitation learners, the relevant conditional variance is the learner's own: it is far below what regression on the demonstrations gives, and it has to be measured from the learner.

\textbf{Noise and robustness in imitation learning.} Training with input noise acts as a regularizer (Bishop 1995). In imitation learning, noise is injected into the demonstrator's actions to widen the state distribution (Laskey et al.~2017), interactive methods correct the covariate shift of behavior cloning (Ross and Bagnell 2010; Ross et al.~2011), and domain randomization perturbs observations for transfer from simulation (Tobin et al.~2017). A control-theoretic line of work explains when imitation errors compound in closed loop. With continuous actions, any smooth, deterministic imitator can suffer execution error exponentially larger, in the horizon, than its error on the expert's data, even when the dynamics are stable (Simchowitz et al.~2025). Matching the expert's higher-order Taylor terms keeps the error small on the learner's own trajectories when the expert is incrementally input-to-state stable (Pfrommer et al.~2022), and injecting noise into the expert's data supervises the directions most prone to compounding (Zhang et al.~2026). These works ask whether noise makes a policy more robust, or how errors in its actions grow, with the state observed exactly. We ask how much a trained policy relies on a noisy perception channel, and predict it.

\textbf{Causal confusion, copycat behavior and shortcuts.} Imitation learners can latch onto nuisance correlates of the expert's action (de Haan et al.~2019), including their own past actions encoded in observation histories (Wen et al.~2020), a case of shortcut learning (Geirhos et al.~2020). The anchor is related but different. The correlation it uses, between robot state and object position, is real in the demonstrations, and the learner weights it roughly in proportion to its in-sample accuracy. It fails in closed loop for a copycat-like reason: the policy itself produces the robot state, so the anchor carries the policy's own past estimates. Recent studies report related failures of proprioceptive inputs in visuomotor policies: reliance on the robot state can cause overfitting to the training trajectories, with poor spatial generalization (Zhao et al.~2025), and can suppress visual learning in the motion-transition phases where the robot must localize its target (Lu et al.~2026).

\textbf{Diffusion policies and action chunking.} We study Diffusion Policy (Chi et al.~2023) and an MLP that predicts the same action chunks. Executing predicted chunks open loop provably avoids exponential compounding of imitation error when the open-loop dynamics are stable (Zhang et al.~2026). With chunked execution (Zhao et al.~2023), clean and noise-trained policies alike also average fast reading noise over several queries (Section~\ref{sec:closed}).

\textbf{Cue combination.} The anchor law is the reliability-weighted cue combination of perception science: a calibrated observer weights each cue by its inverse variance (Ernst and Banks 2002; Körding and Wolpert 2004). Our learners behave roughly like such observers, calibrated in-sample to their own training distribution.

\textbf{Pre-registration.} Writing predictions and decision rules down before the data separates prediction from postdiction (Nosek et al.~2018), and has been proposed for machine learning research as an alternative publication model (Bertinetto et al.~2021). We use it for every test in this paper, including the ones that failed.

\hypertarget{theory}{%
\section{Theory}\label{theory}}

A policy trained on noisy readings responds to the reading in proportion to what the reading adds to what its clean inputs already say about the object. This section derives that weight, a bias identity that tells an anchor from noise inflation, and what changes in closed loop.

\hypertarget{setup}{%
\subsection{Setup}\label{setup}}

Let \(W\) be the object position, \(Y = W + \varepsilon\) the reading with \(\varepsilon \sim \mathcal{N}(0, s^2 I)\) independent across time steps, and \(C\) the learner\textquotesingle s clean inputs (end-effector position and orientation, gripper opening, and the object's orientation, which is not noised). The demonstrator acts as \(a = \pi^*(C, W)\). A learner trained with squared error on \((C, Y) \to a\) approximates \(\mathbb{E}[a \mid C, Y]\). A diffusion policy models the conditional distribution of \(a\), and its mean action obeys the same relation.

Linearize \(\pi^*\) in \(W\) around the anchor \(m(C) = \mathbb{E}[W \mid C]\), with \(G = \partial \pi^*/\partial W\). If \(W\) given \(C\) is Gaussian with covariance \(V\), the learner's implied object estimate and its response to the reading are

\[
\hat{W}(C, Y) = m(C) + \Lambda\,\bigl(Y - m(C)\bigr), \qquad \Lambda = V\,(V + s^2 I)^{-1}, \qquad \frac{\partial a}{\partial Y} = G\,\Lambda
\]

The clean policy (\(s = 0\)) has sensitivity \(G\), so the gain ratio on axis \(j\) is \(\alpha_j = V_j/(V_j + s^2)\) for diagonal \(V\). With an \(n\)-frame history of independent readings, \(s^2\) becomes \(s^2/n\).

\hypertarget{classical-eiv-and-the-anchor-law}{%
\subsection{Classical EIV and the anchor law}\label{classical-eiv-and-the-anchor-law}}

The two predictions differ only in where \(V\) comes from.

\begin{itemize}
\tightlist
\item
  \textbf{Classical EIV} takes \(V\) from a model of the demonstrations. With no informative clean inputs, \(V\) is the population spread \(\mathrm{Var}(W)\). The benchmark used the standard practical version: a linear-Gaussian model of \(W\) given end-effector position and gripper opening, fitted within 1-s time bins, whose residual covariance is \(V\). In the toy tasks of Appendix~\ref{app:toy} the only clean input is the task phase, so \(V = \mathrm{Var}(W \mid \text{phase})\) and the classical prediction is complete.
\item
  \textbf{The anchor law} takes \(V\) from the learner. A noise-trained policy's measured ratio \(\alpha\) at a calibration level \(s_c\) fixes \(V\) for each informative axis \(j\) and probe time \(t\), and \(V\) then predicts the ratio at any other level:
\end{itemize}

\[
V_{jt} = s_c^2\,\frac{\alpha_{jt}(s_c)}{1 - \alpha_{jt}(s_c)}, \qquad \alpha_{jt}(s) = \frac{V_{jt}}{V_{jt} + s^2}
\]

Equivalently, the odds \(\alpha/(1 - \alpha)\) scale as \(s^{-2}\). Nothing is fitted at the new level: \(V\) comes from the calibration policy and the exponent from the Gaussian derivation. The one assumption is that \(V\) is a property of the learner and the data, not of the noise level.

\hypertarget{the-bias-identity}{%
\subsection{The bias identity}\label{the-bias-identity}}

A learner that acts as if the noise were larger than it is, with no anchor, produces the same sensitivity curve. The bias at clean readings separates the two. With \(Y = W\), the implied estimate errs by \(-(I - \Lambda)(W - m(C))\). Regressing this error on \(W - \mu\) across states, with \(\mu\) the population mean, gives \(b_j = -(1 - \alpha_j)\,\mathrm{Cov}(W_j - m_j, W_j)/\mathrm{Var}(W_j)\).

If the anchor is the conditional mean, \(\mathrm{Cov}(W - m, W) = V\), so the bias is small whenever the anchor is informative. Under noise inflation the learner shrinks toward \(\mu\), so \(m = \mu\) and the bias is large:

\[
b_j^{\text{anchor}} = -\frac{\alpha_j\, s^2}{\operatorname{Var}(W_j)}, \qquad b_j^{\text{inflation}} = -(1 - \alpha_j)
\]

The anchor identity has no free parameter: \(\alpha\) is measured and \(\mathrm{Var}(W)\) is the population spread of the demonstrations. It is not the conditional variance, which would count the robot-state information twice (Section~\ref{sec:bias}). The test is stronger in one direction than the other. A bias near \(-(1 - \alpha)\) would show noise inflation, and a small bias rules it out. But a small bias does not by itself show that the anchor is the conditional mean, because an anchor whose error is uncorrelated with \(W\) has no bias at all.

\hypertarget{the-closed-loop-identity}{%
\subsection{The closed-loop identity}\label{the-closed-loop-identity}}

In closed loop the robot state is produced by the policy, starting from a reset. Suppose the reading carries a persistent offset, \(Y = W + \delta\) throughout the episode. Until the gripper touches the object, the policy's inputs are exactly those it would receive with clean readings of an object displaced by \(\delta\): the same readings, and the robot states it produced from them.

So the closed-loop response to a persistent offset equals the response to a real displacement, the policy's tracking gain, which is about 1 for a competent policy whatever \(\alpha\) is. A response below 1 then means the policy also under-tracks the real object. Fast noise is different: independent readings can be averaged over queries, by an anchor that carries past readings forward or by chunked execution.

\hypertarget{second-order-pooling-across-states}{%
\subsection{Second order: pooling across states}\label{second-order-pooling-across-states}}

The registered predictions use one \(V\) per (axis, time) cell, pooled over the probe states in that cell. If \(V\) varies across states, pooling biases the prediction. The map from the calibration weight to the weight at level \(s\),

\[
h(\alpha) = \frac{\alpha\, s_c^2}{\alpha\, s_c^2 + (1 - \alpha)\, s^2}
\]

is convex for \(s > s_c\) and concave for \(s < s_c\). By Jensen's inequality, the pooled prediction is too low at higher noise and too high at lower noise. This predicts the sign of the residuals, although the argument was developed after the first residuals had been seen (Appendix~\ref{app:second}).

\hypertarget{protocol}{%
\section{Protocol}\label{protocol}}

All robot experiments use two robomimic tasks, two learners, synthetic Gaussian noise on the object position, and one probe of the trained policy's sensitivity.

\hypertarget{tasks-and-data}{%
\subsection{Tasks and data}\label{tasks-and-data}}

We use the robomimic Square (nut assembly) and Can (pick and place) tasks with their proficient-human datasets: 200 demonstrations each, low-dimensional observations (Mandlekar et al.~2021), simulated in robosuite 1.4.1 on MuJoCo 3.2.6 (Zhu et al.~2020; Todorov et al.~2012). The 23-dimensional observation holds the object's position and orientation, both also relative to the end effector, plus the end-effector position and orientation and the two gripper finger positions. Actions are the datasets' 7-dimensional end-effector commands.

An axis is informative if the object's initial position varies along it. On Square the nut varies along y (population standard deviation 33.4 mm; x 1.4 mm, z 0). On Can the can varies along x and y (71.0 and 100.3 mm), and a registered rule (standard deviation above 10 mm) selected both.

\hypertarget{noise-model}{%
\subsection{Noise model}\label{noise-model}}

Independent Gaussian noise with the same standard deviation on x, y and z is added to the object position in every training observation. The relative position is recomputed from the clean end-effector pose (geometric error below \(2 \times 10^{-7}\,\mathrm{m}\)), and the actions are untouched. Draws are seeded per demonstration and shared across noise levels, rescaled.

The calibration level is 11.99 mm on Square, the benchmark's level, chosen from the demonstrations before any noise-trained policy existed. On Can it is 12.01 mm, set by a registered rule (classical EIV's mean shrinkage of 0.85 on the informative axes). Test levels are twice and half the calibration level, plus 0.75 and 1.5 times it on Can (9.01 and 18.01 mm). The benchmark also used a 1.13-mm level, AR(1) noise with per-step correlation 0.78 (time constant 0.2 s), and histories of 1, 2 and 4 frames.

\hypertarget{learners}{%
\subsection{Learners}\label{learners}}

The diffusion policy follows Chi et al.~(2023): a temporal U-Net with channel widths 128, 256 and 512, a cosine noise schedule with 100 steps (Nichol and Dhariwal 2021), noise prediction (Ho et al.~2020), and 10 DDIM sampling steps at inference (Song et al.~2021). It predicts 16 actions and executes 8 per query. The MLP has four hidden layers of 512 units and predicts the same 16-action chunk with squared error.

Both learners use AdamW (Loshchilov and Hutter 2019; learning rate \(10^{-4}\)), batch size 256, 40,000 iterations with a cosine schedule, and an exponential moving average of the weights (0.999). Observations are single frames except in the benchmark's history conditions. Every configuration is trained with seeds 0 and 1, the diffusion policy on one T4 GPU and most MLPs on CPU.

\hypertarget{the-gain-ratio-probe}{%
\subsection{The gain-ratio probe}\label{the-gain-ratio-probe}}

The probe states are the first 40 demonstrations at time steps 10, 15, 20, 25 and 30, which is 0.5 to 1.5 s into the episode and before any grasp. At each state we measure the Jacobian of the mean executed translational action (the 8 executed steps) with respect to the observed object position. We use central differences of ±5 mm, applied in all history frames with the relative position recomputed. For the diffusion policy the mean is over 8 sampling draws, with the same initial noise for both perturbations (common random numbers).

The gain ratio compares the noise-trained policy's Jacobian with that of the clean policy of the same seed, as a clean-Jacobian-weighted projection over the informative columns:

\[
\alpha = \frac{\sum_{p,i,j} J^{\text{noisy}}_{p,ij}\, J^{\text{clean}}_{p,ij}}{\sum_{p,i,j} \bigl(J^{\text{clean}}_{p,ij}\bigr)^2}
\]

The sum runs over probe states p where the clean policy's executed action stays below 0.8 in normalized units, since clipping flattens measured sensitivity. Each test registered its gates with its rule, from this set: exact noise geometry, clean Jacobians reproducible across seeds (cosine at least 0.5), at least 25\% unsaturated states, and clean closed-loop success. A competence gate guards a ratio near zero against a policy that simply failed.

\hypertarget{closed-loop-measurements}{%
\subsection{Closed-loop measurements}\label{closed-loop-measurements}}

Closed-loop tests share episode seeds across all policies and conditions, with a 400-step horizon. Conditions are clean readings, independent reading noise at the calibration level, a persistent reading offset of ±10 mm on each informative axis, and a large persistent offset (20 mm on Square, 40 mm on Can, set by a registered pilot rule). The aim is the end-effector position at the first gripper-close command minus the true object position. Closed-loop sensitivity \(\rho\) is the change in aim per unit offset, relative to the clean policy of the same seed. Success is the environment's own check.

\hypertarget{pre-registration}{%
\subsection{Pre-registration}\label{pre-registration}}

For every test, the question, design, decision rule and predicted numbers were written to a timestamped file before the test's data existed. Predictions were computed by fixed code from existing models only. The benchmark's registration records the library's hash, and from D8 on the files record the hash of every script (16-character SHA-256 prefixes). Addenda are timestamped and precede the data they concern. Post hoc analyses are labelled and never enter a verdict.

The last test also has a re-analysis script. It confirms that the prediction file is older than every rollout record, recomputes \(\alpha\) from the raw arrays, and re-runs the registered scoring code. Appendix A lists every registered test with its verdict.

\hypertarget{results}{%
\section{Results}\label{results}}

We report four results, in the order of the contributions. The anchor law predicts a policy's sensitivity to the reading at new noise levels (Section~\ref{sec:law}). Classical EIV misses because the learner's anchor is far sharper than any model of the demonstrations suggests (Section~\ref{sec:eiv}). The bias at clean readings rules out the alternative that the policies merely act as if the noise were larger (Section~\ref{sec:bias}). In closed loop the protection is lost, and open-loop $\alpha$ predicts success (Section~\ref{sec:closed}).

Every result below comes from a test whose rule was registered before its data. Analyses that were not part of a registered rule are marked post hoc. Appendix~\ref{the-pre-registration-record} lists each test with its rule and verdict. Appendices~\ref{app:toy} and~\ref{app:second} report the toy tasks and the second-order deviations.

\subsection{The anchor law predicts sensitivity at new noise levels}\label{sec:law}

\begin{figure}[t]
\centering
\includegraphics[width=0.5\linewidth]{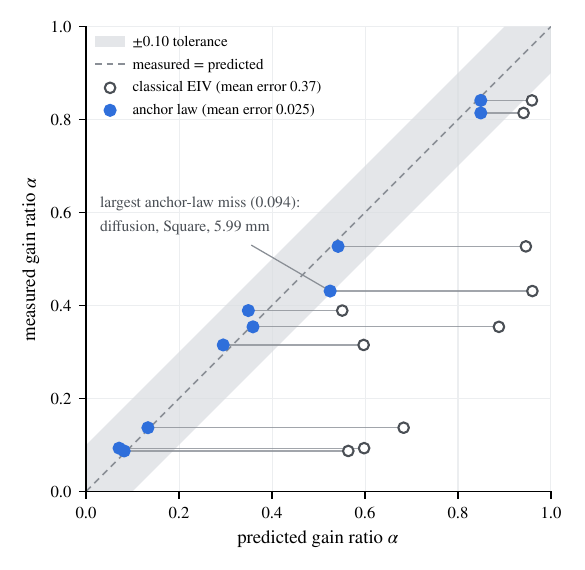}
\caption{The ten prospective predictions of Table~\ref{tab:ten}. Filled: the anchor law, with $V$ measured on the 12-mm policy of the same seed. Open: classical EIV. A horizontal segment joins the two predictions for one measurement. Dashed: measured equals predicted; shaded: $\pm 0.10$. Means of two seeds.}
\label{fig:pred}
\end{figure}

With $V$ measured once, on the 12-mm policy of each seed, the anchor law predicted the gain ratio at all ten held-out noise levels within the registered tolerance. Classical EIV was outside the tolerance at all ten (Table~\ref{tab:ten}, Figure~\ref{fig:pred}).

The tests cover both learners on both tasks at twice and half the calibration level, plus diffusion on Can at 0.75 and 1.5 times it. For every informative axis and probe time, $V$ came from the 12-mm policy of the same seed, and fixed code wrote the predictions to file before any test policy was trained. The tolerance was 0.10, except for one registered secondary, the MLP on Square at 6\,mm, with 0.05. The diffusion test on Square was registered before the MLP's result was known. The first Can test was registered before the Can dataset had been downloaded, and its axes, probe times and calibration level came from registered rules.

\begin{table}[t]
\centering
\caption{Gain ratio $\alpha$ on the informative axes at held-out noise levels, mean of two seeds. Both predictions use $\alpha = V/(V + s^2)$. The anchor law takes $V$ from the 12-mm policy of the same learner and seed; classical EIV takes it from a linear model of the demonstrations. Every prediction was written to file before the test policies were trained. Tolerance 0.10 (0.05 for the MLP on Square at 6\,mm).}
\label{tab:ten}
\small
\begin{tabular}{@{}llrrrr@{}}
\toprule
Task & Learner & Noise $s$ (mm) & Measured & Anchor law & Classical EIV \\
\midrule
Square & MLP & 24 & 0.315 & 0.295 & 0.597 \\
 & & 6 & 0.841 & 0.849 & 0.959 \\
Square & diffusion & 24 & 0.093 & 0.071 & 0.598 \\
 & & 6 & 0.431 & 0.525 & 0.960 \\
Can & MLP & 24 & 0.389 & 0.349 & 0.551 \\
 & & 6 & 0.814 & 0.849 & 0.941 \\
Can & diffusion & 24 & 0.087 & 0.082 & 0.564 \\
 & & 6 & 0.527 & 0.542 & 0.946 \\
 & & 9 & 0.354 & 0.359 & 0.888 \\
 & & 18 & 0.137 & 0.133 & 0.683 \\
\midrule
\multicolumn{4}{@{}l}{Mean absolute error} & 0.025 & 0.37 \\
\multicolumn{4}{@{}l}{Within tolerance} & 10 of 10 & 0 of 10 \\
\bottomrule
\end{tabular}
\end{table}

The anchor law's mean absolute error over the ten predictions is 0.025, and classical EIV's is 0.37. The smallest classical error, 0.12, is larger than the largest anchor-law error. That largest error is diffusion on Square at 6\,mm (measured 0.431, predicted 0.525), inside the tolerance only at its edge (Appendix~\ref{app:second}). The two seeds agreed closely: within 0.01 for diffusion on Can at 6 and 24\,mm, and within 0.02 at 9 and 18\,mm.

The predictions are learner-specific, which classical EIV's cannot be. On Square at 24\,mm, classical EIV gives 0.597 and 0.598 for the two learners, because it depends only on the data. The measured ratios differ by a factor of three (MLP 0.315, diffusion 0.093), and the anchor law, calibrated on each learner, predicted both (0.295 and 0.071).

\subsection{Why classical EIV misses: the learner's anchor is sharper than the data's}\label{sec:eiv}

Classical EIV was the first prediction we registered, and it failed. For diffusion policies on Square we registered it for seven noise conditions, two seeds each, before any noise-trained policy existed (Table~\ref{tab:bench}). The rule required a mean absolute error of at most 0.15 and a Spearman correlation of at least 0.7. The rank order held (Spearman 0.85), but the mean absolute error was 0.245. A companion prediction for longer histories failed too: four frames instead of one raised the ratio by 0.30 under iid noise, against a predicted 0.07, although under AR(1) noise the gain (0.13 against 0.00) stayed within tolerance. On the informative axis the miss is large: the single-frame 12-mm policies kept 23\% of their clean sensitivity, where classical EIV predicted 86\%.

\begin{table}[t]
\centering
\caption{The registered classical prediction for diffusion policies on Square: gain ratio over all three object axes, mean of two seeds.}
\label{tab:bench}
\small
\begin{tabular}{@{}lrrr@{}}
\toprule
Condition & History (frames) & Classical EIV & Measured \\
\midrule
iid, 1.13\,mm & 2 & 0.874 & 0.941 \\
iid, 11.99\,mm & 1 & 0.584 & 0.187 \\
iid, 11.99\,mm & 2 & 0.624 & 0.309 \\
iid, 11.99\,mm & 4 & 0.652 & 0.490 \\
AR(1), 11.99\,mm & 1 & 0.584 & 0.273 \\
AR(1), 11.99\,mm & 2 & 0.582 & 0.303 \\
AR(1), 11.99\,mm & 4 & 0.588 & 0.407 \\
\bottomrule
\end{tabular}
\end{table}

The two predictions share the formula $\alpha = V/(V + s^2)$ and differ only in $V$. Classical EIV takes $V$ from a model of the demonstrations. Every linear and nearest-neighbour model of the nut position given the clean features puts the anchor's error near 30\,mm and predicts the same ratios (mean absolute error 0.23 to 0.25). The trained policies act as if it were much smaller. In the pooled approximation the anchor's error is about 7\,mm for diffusion (6.5 on Square, 6.9 on Can) and about 15\,mm for the MLP (14.7 and 15.6), which is why diffusion policies discount the reading more. Four registered tests ruled out the explanations that would make this gap an artefact (Table~\ref{tab:alt}).

\begin{table}[t]
\centering
\caption{Explanations of the classical miss that registered tests ruled out. Bias values are at clean readings, in units of the clean sensitivity, for diffusion on Square. For the MLP the recall test was intermediate ($-0.05$, $-0.07$ and $-0.12$, against $-0.37$ to $-0.34$ for plain shrinkage).}
\label{tab:alt}
\small
\begin{tabular}{@{}p{0.2\linewidth}p{0.335\linewidth}p{0.39\linewidth}@{}}
\toprule
Explanation & What it predicts & What we found \\
\midrule
A mis-estimated spread & A better model of the demonstrations fixes the prediction & Every linear and nearest-neighbour model misses alike (mean absolute error 0.23 to 0.25) \\
\addlinespace
Local memorization & A stronger discount at training states than at fresh placements & The same discount at both (difference $+0.02$); the nut's orientation, unique to each demonstration, is not used as a lookup \\
\addlinespace
Noise structure within a demonstration (MLP) & Noise held constant per demonstration is discounted much less & It is not: 0.660, against 0.602 for independent frames and 0.633 for 1-s correlated noise (the rule needed $+0.15$) \\
\addlinespace
Recall of training placements & No bias at training placements, a bias at fresh ones & Near-zero bias at both: $-0.008$ at demonstration states, $-0.029$ and $-0.033$ along trajectories from training and fresh placements, against $-0.64$ for plain shrinkage \\
\bottomrule
\end{tabular}
\end{table}

Two further tests characterize the anchor. It depends on the amount of data: trained on 50, 100 and 200 demonstrations, the 12-mm MLP's ratio was 0.362, 0.500 and 0.602. With less data, the anchor fits the training set more tightly than it generalizes, and the learner trusts it more. And its weighting roughly matches its in-sample accuracy. Let $\kappa$ be the anchor's actual error divided by the error its weighting implies. For diffusion, $\kappa$ was 1.03 at training states and 1.27 at fresh placements, inside the registered bound of 1.3 for a calibrated anchor. The margin is thin: without two clipped estimates the fresh value is 1.52, inside the registered overconfident band, and a robust version gives 1.25. A registered check that $\kappa$ at training states stays near 1 at other noise levels held at 24\,mm (1.12) but not at 6\,mm (0.08), where $\alpha$ is 0.84 and the estimator, which divides by $1 - \alpha$, is probably unreliable. For the MLP with 200 demonstrations, $\kappa$ at fresh placements was 1.1 to 1.4, rising to 1.7 to 3.5 with 50 demonstrations (post hoc).

So the learners act roughly as reliability-weighted cue combiners: the weighting matches the anchor's in-sample accuracy and is mildly overconfident at new placements. Held-out regressors give only a lower bound on how much the robot state reveals: a single neural regressor recovers less of the nut position from the clean frame ($R^2$ 0.54 at $t = 30$) than an ensemble does. Where the clean inputs carry no information about the object, the two predictions coincide, and classical EIV is then accurate: on two toy tasks whose only clean input is the task phase, it matched the learned gain within 0.13 on a held-out task (Appendix~\ref{app:toy}).

\subsection{The bias at clean readings rules out noise inflation}\label{sec:bias}

An anchor and a learner that acts as if the noise were larger produce the same sensitivity curve, and the bias at clean readings tells them apart (Section~\ref{the-bias-identity}). For both learners on both tasks it is small, $-0.003$ to $-0.052$, where noise inflation predicts $-0.37$ to $-0.75$ (Table~\ref{tab:bias}). On Square the bias also stays small along trajectories that a clean policy drives from training and from fresh placements (diffusion $-0.029$ and $-0.033$; MLP $-0.07$ and $-0.12$).

\begin{table}[t]
\centering
\caption{Bias at clean readings at 12\,mm, measured at demonstration states, in units of the clean sensitivity. Square rows average over the 12-mm conditions of earlier tests (noise structures, and for diffusion also history lengths); Can rows are single-frame iid. Anchor predictions: $^a$the registered formula, with the conditional variance; $^b$the corrected formula, with the population variance, averaged over the same conditions as the measurement.}
\label{tab:bias}
\small
\begin{tabular}{@{}llp{0.17\linewidth}p{0.3\linewidth}l@{}}
\toprule
Task & Learner & Measured bias & Anchor prediction & Noise inflation \\
\midrule
Square & diffusion & $-0.008$ & $-0.041^a$; $-0.031^b$ (post hoc) & $-0.64$ \\
Square & MLP & $-0.052$ & $-0.082^b$ (post hoc) & $-0.37$ \\
Can & MLP & $-0.008$ & $-0.111^a$; within 0.008 per axis$^b$ (post hoc) & $-0.373$ \\
Can & diffusion & $x$ $-0.004$, $y$ $-0.003$ & $x$ $-0.007$, $y$ $-0.003^b$ & $-0.752$ \\
\bottomrule
\end{tabular}
\end{table}

The anchor predictions carry a correction. The registered formulas used the conditional variance of the object position where the identity needs the population variance, which counts the robot-state information twice. On Can this mattered (population 71 and 100\,mm, conditional 33 and 25\,mm), and the MLP's registered verdict there is intermediate as a result. The corrected formula was found on the MLP (post hoc), where it matches each axis within 0.008 at all three noise levels. It was then registered before the diffusion data, and it held. For diffusion it predicts a bias close to zero, so that test separates it from the conditional formula ($x$ $-0.034$, $y$ $-0.054$) and from noise inflation, but not from zero.

On Square the correction makes little difference (nut $y$: population 33.4\,mm, conditional 29.5\,mm), and the corrected formula was computed only after the data (post hoc). For the MLP it gives $-0.082$, averaged over the same six policies as the measurement. The measured $-0.052$ has a 95\% interval of $-0.103$ to $-0.007$ from a bootstrap over the 40 probe demonstrations, which contains the prediction and excludes zero. For diffusion the formula gives $-0.031$ ($-0.016$ to $-0.041$ over the six 12-mm conditions), against $-0.008$ measured, and we have no interval for that test. So the size of the bias agrees with a conditional-mean anchor on Can and, within a wide interval, for the MLP on Square, but for diffusion on Square it is smaller. As Section~\ref{the-bias-identity} notes, a small bias does not by itself show that the anchor is the conditional mean. What holds on all four rows is the exclusion of noise inflation. Together with the calibrated weighting (Section~\ref{sec:eiv}), this points to a learned anchor rather than a mis-scaled noise model.

\subsection{In closed loop the protection is lost, and \texorpdfstring{$\alpha$}{alpha} predicts success}\label{sec:closed}

\begin{figure}[t]
\centering
\includegraphics[width=\linewidth]{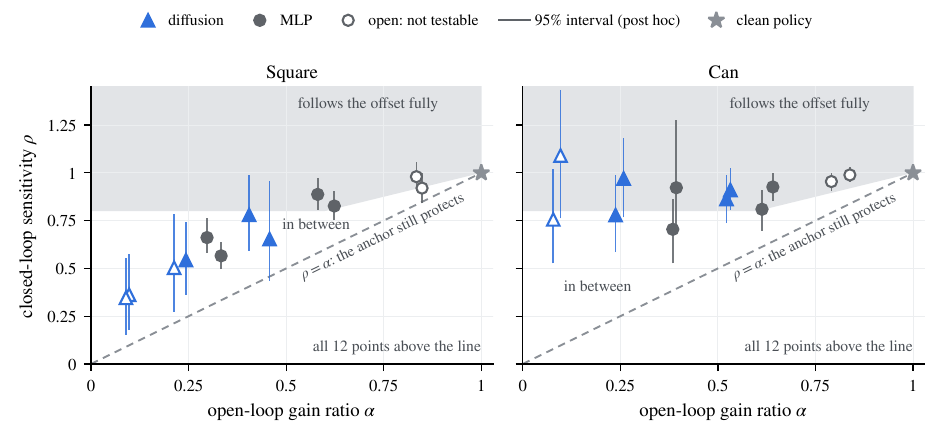}
\caption{Closed-loop sensitivity $\rho$ to a persistent $\pm 10$\,mm reading offset against open-loop $\alpha$ for all 24 noise-trained policies: two learners on two tasks, three noise levels, two seeds. Dashed: $\rho = \alpha$, where the anchor would still protect. Shaded: the registered criterion for following the offset fully ($\rho \ge 0.80$ and above $(1 + \alpha)/2$); between it and the dashed line, in between. Filled: the 15 policies that passed the competence gate and whose $\alpha$ was far enough from 1 to separate the predictions; open: the other 9. Lines: 95\% intervals for $\rho$ from a bootstrap over episode placements, each policy paired with the clean policy of the same seed (post hoc; $\alpha$ held at its measured value). On Can, the two seeds of diffusion at 12\,mm and at 24\,mm are spread by less than 0.01 in $\alpha$ for legibility. Every point lies above the dashed line, and 22 of the 24 intervals lie entirely above it.}
\label{fig:rho}
\end{figure}

In closed loop no policy kept its open-loop protection against a persistent reading error. Each learner was rolled out on both tasks as a clean policy and as policies trained at 6, 12 and 24\,mm, two seeds each. The registered alternative to the closed-loop identity (Section~\ref{the-closed-loop-identity}) was that the anchor keeps protecting, so that the sensitivity $\rho$ to a persistent $\pm 10$\,mm offset stays near $\alpha$. It did not: the point estimate of $\rho$ exceeded $\alpha$ in all 24 noise-trained policies, by 0.20 to 0.53 for the eight discriminating MLP policies, and the 95\% intervals of 22 of them lie entirely above $\alpha$ (post hoc; Figure~\ref{fig:rho}). Of the 15 policies that passed the competence gate and could separate the two readings, 8 followed the offset fully ($\rho \ge 0.80$ and above $(1 + \alpha)/2$), 7 were in between and none kept its open-loop $\alpha$, although for one of them, diffusion on Square at 6\,mm, the interval reaches down to $\alpha$. The registered prediction was stricter, that every policy follows the offset fully, so the test's verdict is mixed for both learners.

\begin{figure}[t]
\centering
\includegraphics[width=\linewidth]{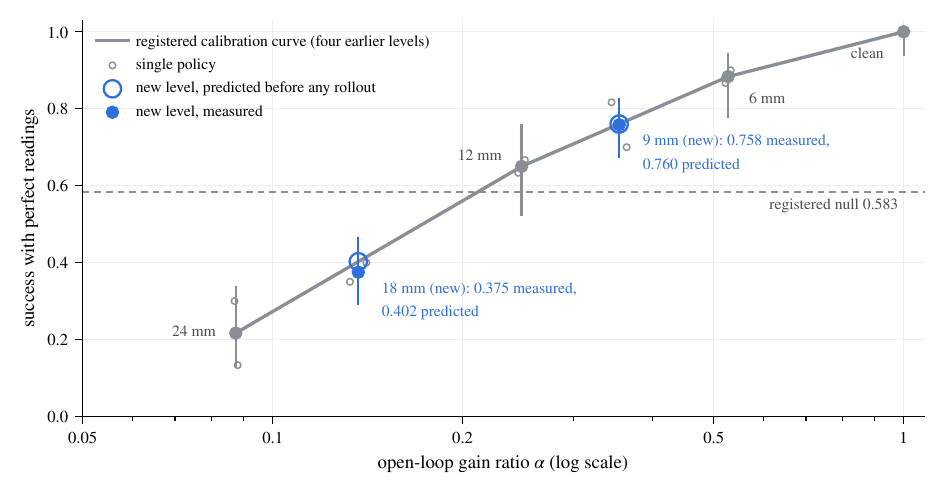}
\caption{Closed-loop success with perfect readings against open-loop $\alpha$ for diffusion policies on Can. Grey: the four levels of the closed-loop offset test, which fix the registered calibration curve, piecewise linear in $\ln\alpha$. Blue: the two new levels of the prospective test, measured (filled) and predicted from the measured $\alpha$ before any rollout (rings). Small circles: single policies. Whiskers: 95\% Wilson intervals. Dashed: the registered null, the mean success of the six earlier noise-trained policies.}
\label{fig:success}
\end{figure}

For the MLP it is clear what the in-between policies are doing (post hoc). Moving the real object by $\pm 10$\,mm, with clean readings, moves their aim almost exactly as much as the reading offset does: $r = 0.94$ across all 16 MLP policies. The real displacement moved the aim slightly more in all 16, by 0.06 on average, a small systematic gap beyond the identity. The three in-between MLP policies, all trained at 24\,mm, follow the real object only 60 to 81\% of the way. What survives of the attenuation is sluggishness toward the object, not protection against reading errors. This check has not been run for diffusion, so partial protection in its four in-between policies is not excluded.

Two further registered predictions were also mixed. Under a large offset, noise training raised success over the clean policy by $+0.00$ to $+0.20$, too little to count as protection, on small samples. Under fast noise, the claim that the anchor adds temporal averaging was not supported: the clean policies themselves weight readings two to five queries back (0.6 to 0.8 for the MLP, 0.55 to 0.93 for diffusion), plausibly because chunked execution averages them.

The cost shows in success (post hoc). With perfect readings, the diffusion policies' success falls steeply as their open-loop $\alpha$ falls (Spearman 0.93 over 16 policies). On Can it is 1.00, 0.88, 0.65 and 0.22 for clean, 6, 12 and 24\,mm, and on Square 0.82, 0.70, 0.43 and 0.43. For the MLP the effect is weak (Spearman 0.34). Noise training did not buy diffusion policies robustness to fast noise either. Under 12-mm reading noise, no noise-trained diffusion configuration beat the clean policy on average (Square 0.00 to $-0.10$, Can $-0.12$ to $-0.72$), while noise-trained MLPs did on Square ($+0.17$ to $+0.30$). The policy that looks most robust in open loop, diffusion with the lowest $\alpha$, is the most fragile in closed loop.

We then tested the relation prospectively. We trained diffusion policies on Can at two new noise levels, 9.01 and 18.01\,mm, and registered the prediction rules before training. The anchor law predicted their $\alpha$ from the 12-mm calibration. A calibration curve, piecewise linear in $\ln\alpha$ through the four earlier levels, predicted success from the measured $\alpha$; those predictions were written after training and probing, before any rollout. The null was the mean success of the six earlier noise-trained policies, 0.583. Confirmation required both levels within 0.15 of the prediction and a smaller summed error than the null. A validity rule required each level's $\alpha$ to fall between its neighbours', so that a failed run with $\alpha$ near zero could not confirm a low predicted success. Each policy ran 60 episodes with perfect readings, the first 30 on the earlier placements, and 30 under 12-mm reading noise.

All four registered predictions, the primary and three secondaries, were confirmed (Figure~\ref{fig:success}), and the re-analysis script reproduced them from the raw arrays. The measured $\alpha$ was 0.354 and 0.137 against the law's 0.359 and 0.133 (classical EIV 0.888 and 0.683). Success with perfect readings was 0.758 (91 of 120) and 0.375 (45 of 120), against 0.760 and 0.402 predicted from the measured $\alpha$. The full chain, from the law's $\alpha$ with no new probes, predicted 0.765 and 0.393. Under 12-mm reading noise, success was 0.733 and 0.450 against 0.708 and 0.395. Across all 12 diffusion policies on Can, Spearman($\alpha$, success) is 0.986 (post hoc).

The registered scope is narrower than ``$\alpha$ causes failure''. At these levels $\alpha$ and training noise move together, and a curve interpolated in training noise predicts success as well (0.747 and 0.397; post hoc). The test establishes open-loop $\alpha$ as a predictor of closed-loop success for diffusion on Can: given a curve calibrated on other policies' rollouts, it predicts a new policy's success without rolling that policy out, whether $\alpha$ is measured or computed from the law. It does not separate $\alpha$ from the training noise as the cause.

\hypertarget{limitations-and-open-questions}{%
\section{Limitations and open questions}\label{limitations-and-open-questions}}

The main results are prospective and registered, but their scope is narrow in seven ways.

\begin{itemize}
\tightlist
\item
  \textbf{Simulation with synthetic noise.} All noise is Gaussian, added to the object position in a low-dimensional state, and independent across time steps except in the benchmark's AR(1) conditions. Real perception errors are structured: they depend on viewpoint, occlusion and object pose, they can be heavy-tailed, and they arise inside image-based policies. Whether the law holds for visuomotor policies is untested.
\item
  \textbf{\(V\) is measured, not derived.} The law predicts across noise levels from one noise-trained policy. It does not predict a policy's discount from the data alone, because standard estimators of the demonstrations put \(V\) several times too high, and \(V\) also depends on the amount of data (Section~\ref{sec:eiv}).
\item
  \textbf{Uninformative axes.} On Square's x axis, where the nut's position varies by only 1.4 mm, the diffusion policies keep about as much sensitivity as on y (0.37 against 0.38, averaged post hoc over 36 policy-conditions). A calibrated anchor would allow at most 0.05 there. The law and the calibration claims were tested on informative axes only; on an axis with almost no spread the learners are not Bayes-optimal.
\item
  \textbf{Two seeds per configuration.} Seeds agreed closely on sensitivity (within 0.02 for diffusion on Can), but closed-loop success rests on 30 to 60 episodes per policy, and the per-configuration effects in the closed-loop offset test have wide intervals.
\item
  \textbf{\(\alpha\) and training noise are confounded in the prospective success test.} At the tested levels a curve in training noise predicts success as well as the curve in \(\alpha\). Separating them needs policies with different \(\alpha\) at the same training noise, for example from different amounts of data.
\item
  \textbf{Mechanism tests in closed loop were mixed.} The closed-loop offset test's three registered predictions were mixed for both learners, and the real-displacement check that explains the in-between policies was run for the MLP only.
\item
  \textbf{Second-order effects.} The pooling correction (Appendix~\ref{app:second}) is post hoc for the MLP and failed its registered test for diffusion. The diffusion-Square remainder at 6 mm, and the diffusion policies' smaller discount of AR(1) than of iid noise with one frame, are unexplained. The bias identity's size is checked only post hoc for the MLP, where it agrees on both tasks (on Square within a wide interval). For diffusion the registered test on Can cannot separate it from zero, and on Square the diffusion bias is smaller than predicted, with no interval.
\end{itemize}

Two narrower points also bound the claims. The prospective tests were designed to separate the anchor law from classical EIV; they were not designed to separate it from other smooth curves through the calibration point. And sensitivity is probed only on the translational action, at five early time steps before any grasp.

Open questions follow directly. Does the law hold for image-based policies and real perception noise? Can \(V\) be predicted from the data, for example by a held-out regressor matched to the policy's capacity? And can the closed-loop cost be removed, by noise or dropout on the proprioceptive inputs during training, by interactive data collection, or by many more demonstrations?

\hypertarget{conclusion}{%
\section{Conclusion}\label{conclusion}}

An imitation learner trained on noisy object readings does more than discount them by the classical reliability ratio. Because demonstrators steer toward the object, it builds an anchor from its own proprioception and weights the reading by \(\alpha = V/(V + s^2)\), where \(V\) is the anchor\textquotesingle s error variance. Measured once, \(V\) predicted the learner's sensitivity at new noise levels across two tasks and two learner classes, within tolerance in all ten registered predictions.

The same anchor does not keep its protection in deployment. There it reads a robot state the policy produced itself, and none of the 15 policies that could be tested kept its open-loop protection against persistent reading errors, although 7 followed the offset only partly. For diffusion policies, lower \(\alpha\) went with lower success (post hoc), and on Can \(\alpha\) predicted the success of new policies before they were rolled out. That makes \(\alpha\) a candidate diagnostic for noise-trained policies, pending a test that separates it from the training noise.

The path to this law ran through a registered failure. We report the full record, failures included, so that every claim can be traced to the test that produced it.

\section*{Use of AI tools}

The author used Claude (Anthropic) throughout this work. Claude wrote most of the text and code from the author's instructions. The author designed the tests and ran the experiments. The author reviewed all text, code and results and takes full responsibility for the content of this paper.

\hypertarget{references}{%
\section*{References}\label{references}}
\addcontentsline{toc}{section}{References}
{\small

\begin{itemize}
\tightlist
\item
  Bertinetto, L., Henriques, J. F., Albanie, S., Paganini, M., and Varol, G. (Eds.) (2021). \emph{NeurIPS 2020 Workshop on Pre-registration in Machine Learning}. Proceedings of Machine Learning Research, vol.~148.
\item
  Bishop, C. M. (1995). Training with noise is equivalent to Tikhonov regularization. \emph{Neural Computation}, 7(1), 108--116.
\item
  Carroll, R. J., Ruppert, D., Stefanski, L. A., and Crainiceanu, C. M. (2006). \emph{Measurement Error in Nonlinear Models: A Modern Perspective} (2nd ed.). Chapman and Hall/CRC.
\item
  Chi, C., Feng, S., Du, Y., Xu, Z., Cousineau, E., Burchfiel, B., and Song, S. (2023). Diffusion Policy: Visuomotor policy learning via action diffusion. \emph{Robotics: Science and Systems (RSS)}.
\item
  de Haan, P., Jayaraman, D., and Levine, S. (2019). Causal confusion in imitation learning. \emph{Advances in Neural Information Processing Systems (NeurIPS)}.
\item
  Ernst, M. O., and Banks, M. S. (2002). Humans integrate visual and haptic information in a statistically optimal fashion. \emph{Nature}, 415, 429--433.
\item
  Frost, C., and Thompson, S. G. (2000). Correcting for regression dilution bias: comparison of methods for a single predictor variable. \emph{Journal of the Royal Statistical Society: Series A}, 163(2), 173--189.
\item
  Fuller, W. A. (1987). \emph{Measurement Error Models}. Wiley.
\item
  Geirhos, R., Jacobsen, J.-H., Michaelis, C., Zemel, R., Brendel, W., Bethge, M., and Wichmann, F. A. (2020). Shortcut learning in deep neural networks. \emph{Nature Machine Intelligence}, 2, 665--673.
\item
  Ho, J., Jain, A., and Abbeel, P. (2020). Denoising diffusion probabilistic models. \emph{NeurIPS}.
\item
  Körding, K. P., and Wolpert, D. M. (2004). Bayesian integration in sensorimotor learning. \emph{Nature}, 427, 244--247.
\item
  Laskey, M., Lee, J., Fox, R., Dragan, A., and Goldberg, K. (2017). DART: Noise injection for robust imitation learning. \emph{Conference on Robot Learning (CoRL)}.
\item
  Loshchilov, I., and Hutter, F. (2019). Decoupled weight decay regularization. \emph{International Conference on Learning Representations (ICLR)}.
\item
  Lu, J., Xia, W., Wu, Y., Lu, Z., and Hu, D. (2026). When would vision-proprioception policies fail in robotic manipulation? \emph{International Conference on Learning Representations (ICLR)}.
\item
  Mandlekar, A., Xu, D., Wong, J., Nasiriany, S., Wang, C., Kulkarni, R., Fei-Fei, L., Savarese, S., Zhu, Y., and Martín-Martín, R. (2021). What matters in learning from offline human demonstrations for robot manipulation. \emph{CoRL}.
\item
  Nichol, A. Q., and Dhariwal, P. (2021). Improved denoising diffusion probabilistic models. \emph{International Conference on Machine Learning (ICML)}.
\item
  Nosek, B. A., Ebersole, C. R., DeHaven, A. C., and Mellor, D. T. (2018). The preregistration revolution. \emph{Proceedings of the National Academy of Sciences}, 115(11), 2600--2606.
\item
  Pfrommer, D., Zhang, T. T. C. K., Tu, S., and Matni, N. (2022). TaSIL: Taylor series imitation learning. \emph{Advances in Neural Information Processing Systems (NeurIPS)}.
\item
  Ross, S., and Bagnell, J. A. (2010). Efficient reductions for imitation learning. \emph{International Conference on Artificial Intelligence and Statistics (AISTATS)}.
\item
  Ross, S., Gordon, G., and Bagnell, J. A. (2011). A reduction of imitation learning and structured prediction to no-regret online learning. \emph{AISTATS}.
\item
  Simchowitz, M., Pfrommer, D., and Jadbabaie, A. (2025). The pitfalls of imitation learning when actions are continuous. \emph{Conference on Learning Theory (COLT)}.
\item
  Song, J., Meng, C., and Ermon, S. (2021). Denoising diffusion implicit models. \emph{ICLR}.
\item
  Tobin, J., Fong, R., Ray, A., Schneider, J., Zaremba, W., and Abbeel, P. (2017). Domain randomization for transferring deep neural networks from simulation to the real world. \emph{IEEE/RSJ International Conference on Intelligent Robots and Systems (IROS)}.
\item
  Todorov, E., Erez, T., and Tassa, Y. (2012). MuJoCo: A physics engine for model-based control. \emph{IROS}.
\item
  Wen, C., Lin, J., Darrell, T., Jayaraman, D., and Gao, Y. (2020). Fighting copycat agents in behavioral cloning from observation histories. \emph{NeurIPS}.
\item
  Zhang, T. T., Pfrommer, D., Pan, C., Matni, N., and Simchowitz, M. (2026). Action chunking and data augmentation yield exponential improvements in behavior cloning for continuous spaces. \emph{International Conference on Learning Representations (ICLR)}.
\item
  Zhao, J., Lu, W., Zhang, D., Liu, Y., Liang, Y., Zhang, T., Cao, Y., Xie, J., Hu, Y., Wang, S., Guo, J., Wang, D., and Gao, Y. (2025). Do you need proprioceptive states in visuomotor policies? arXiv:2509.18644.
\item
  Zhao, T. Z., Kumar, V., Levine, S., and Finn, C. (2023). Learning fine-grained bimanual manipulation with low-cost hardware. \emph{RSS}.
\item
  Zhu, Y., Wong, J., Mandlekar, A., Martín-Martín, R., et al.~(2020). robosuite: A modular simulation framework and benchmark for robot learning. arXiv:2009.12293.
\end{itemize}
}

\clearpage
\begin{center}
{\LARGE\bfseries Appendix}
\end{center}
\vspace{0.6em}
\appendix
\makeatletter
\renewcommand\@seccntformat[1]{\ifnum\pdfstrcmp{#1}{section}=0 Appendix~\csname the#1\endcsname:\enspace\else\csname the#1\endcsname\quad\fi}
\makeatother

\hypertarget{the-pre-registration-record}{%
\section{The pre-registration record}\label{the-pre-registration-record}}

Every registered test behind this paper, in the order it was written, with its registered verdict. Each file was timestamped before the test's data existed. The two analysis plans were written after the benchmark's outcome was known, so they could only diagnose the miss.

{\footnotesize
\begin{longtable}[]{@{}
  >{\raggedright\arraybackslash}p{(\columnwidth - 8\tabcolsep) * \real{0.1600}}
  >{\raggedright\arraybackslash}p{(\columnwidth - 8\tabcolsep) * \real{0.1200}}
  >{\raggedright\arraybackslash}p{(\columnwidth - 8\tabcolsep) * \real{0.2200}}
  >{\raggedright\arraybackslash}p{(\columnwidth - 8\tabcolsep) * \real{0.2000}}
  >{\raggedright\arraybackslash}p{(\columnwidth - 8\tabcolsep) * \real{0.3000}}@{}}
\toprule\noalign{}
\begin{minipage}[b]{\linewidth}\raggedright
Test
\end{minipage} & \begin{minipage}[b]{\linewidth}\raggedright
Registered (UTC)
\end{minipage} & \begin{minipage}[b]{\linewidth}\raggedright
Question
\end{minipage} & \begin{minipage}[b]{\linewidth}\raggedright
Registered verdict
\end{minipage} & \begin{minipage}[b]{\linewidth}\raggedright
Key numbers
\end{minipage} \\
\midrule\noalign{}
\endhead
\bottomrule\noalign{}
\endlastfoot
PR4, arm & \mbox{2026-09-24} 18:53 & Classical gain on a second toy task & Gain test PASS; overall FAIL & gain within 0.13 at 4 of 4 levels; chunking claim refuted; success ordering 0.74 (needed 0.8) \\
PR5, peg and arm & \mbox{2026-09-24} 19:14 & Success predicted from the classical gain & FAIL & MAE 0.29 over 20 held-out configurations \\
PR6, peg and arm & \mbox{2026-09-24} 22:16 & Same, frozen learner protocol, 14 new configurations & PASS & MAE 0.11 against 0.30 naive \\
PR7 E1 & \mbox{2026-10-01} 01:30 & Does PR6 hold with four seeds? & PR6 STANDS & MAE 0.102 (95\% CI 0.065 to 0.161), naive 0.312, p = 0.0008 \\
PR7 E2 & \mbox{2026-10-01} 01:30 & History length and noise structure & PASS & MAE 0.079, Spearman 0.93 \\
PR7 E3 & \mbox{2026-10-01} 01:30 & Robustness to the binning choice & PASS & MAE change +0.023 and +0.005 (limit 0.05) \\
PR8 & \mbox{2026-10-01} 03:00 & Success under human-like demonstrations & PASS, below the adoption bar & MAE 0.107 against 0.170 naive (bar: 0.10 better) \\
PR9 & \mbox{2026-10-01} 03:55 & Joint-posterior version of PR8 & FAIL & improvement over naive 0.029 (needed 0.03) \\
Benchmark, Square, diffusion & \mbox{2026-10-04} 06:33 & Classical EIV for the gain ratio & FAIL & MAE 0.245 (limit 0.15), Spearman 0.85; iid history gain +0.30 against +0.07 (limit 0.15) \\
Analysis plan: spread & \mbox{2026-10-07} 10:03 & Was the spread mis-estimated? & Not explained by the spread & every estimator MAE 0.23 to 0.25 \\
D1 & \mbox{2026-10-07} 17:44 & Is the extra discount local to training states? & GLOBAL & fresh placements +0.02; no orientation lookup \\
D2 & \mbox{2026-10-07} 19:09 & What kind of deviation? & LINEAR ATTENUATION; MLP comparison INTERMEDIATE & the MLP shows the pattern only partly \\
D3, MLP & \mbox{2026-10-07} 22:19 & Within-demonstration noise contrast? & NULL & 0.660, 0.602, 0.633 (needed +0.15) \\
D4, MLP & \mbox{2026-10-07} 23:19 & Recall of training placements? & INTERMEDIATE & bias −0.05, −0.07, −0.12 against −0.37 to −0.34 \\
D4b, diffusion & \mbox{2026-10-07} 23:34 & Same, main learner & POPULATION-VALID KNOWLEDGE & bias −0.008, −0.029, −0.033 against −0.64 \\
Analysis plan: robot-state information & \mbox{2026-10-08} 00:30 & Is the robot-state information real? & Overweighting by the plan rule, with a caveat & held-out \(R^2\) 0.54 at t = 30, a lower bound \\
D5, MLP & \mbox{2026-10-08} 01:00 & Finite-sample effect? & FINITE-SAMPLE OVERCONFIDENCE & 0.362, 0.500, 0.602 at 50, 100, 200 demonstrations; 100 inside an illustrative 0.50 to 0.55 \\
D4c, diffusion & \mbox{2026-10-08} 09:31 & Is the anchor's weighting calibrated? & CALIBRATED, at the margin & \(\kappa\) 1.03 at training states, 1.27 at fresh placements (limit 1.3); 1.52 without two clipped estimates \\
D6, MLP, Square & \mbox{2026-10-08} 22:22 & Anchor law at new noise levels & ANCHOR MODEL CONFIRMED & 0.315 vs 0.295 at 24 mm; 0.841 vs 0.849 at 6 mm (secondary); \(\kappa\) at training states 1.12 and 0.08 (registered: about 1) \\
D6b, diffusion, Square & \mbox{2026-10-08} 22:24 & Same, main learner & ANCHOR MODEL CONFIRMED & 0.093 vs 0.071; 0.431 vs 0.525 (secondary, at the edge) \\
D7, MLP, Can & \mbox{2026-10-09} 03:42 & Anchor law on a second task & ANCHOR LAW CONFIRMED & 0.389 vs 0.349; 0.814 vs 0.849 \\
D7b, diffusion, Can & \mbox{2026-10-09} 09:16 & Same, main learner & ANCHOR LAW CONFIRMED & 0.087 vs 0.082; 0.527 vs 0.542 \\
D8 part A, MLP, Can & \mbox{2026-10-09} 17:56 & Anchor or noise inflation? & INTERMEDIATE; noise inflation excluded & bias −0.008 against −0.111 (formula) and −0.373 (inflation) \\
D8 part B, diffusion & \mbox{2026-10-09} 17:56 & Does pooling explain the residuals? & NOT EXPLAINED & Square residual cut 48\% (needed 50\%); Can overshoot at 6 mm \\
D8 part A, diffusion, Can & \mbox{2026-10-09} 17:59 & Same, corrected formula & ANCHOR; corrected formula confirmed & bias −0.004 against −0.752 (inflation) \\
D9, MLP & \mbox{2026-10-09} 19:11 & Does the protection survive closed loop? & MIXED (P1, P2, P3) & 5 follow fully, 3 in between, 0 protect \\
D9, diffusion & \mbox{2026-10-09} 19:11 & Same, main learner & MIXED (P1, P2, P3) & 3 follow fully, 4 in between, 0 protect \\
D10, diffusion, Can & \mbox{2026-10-10} 03:51 & Does open-loop \(\alpha\) predict closed-loop success? & CONFIRMED (primary and three secondaries) & success 0.758 vs 0.760 and 0.375 vs 0.402; \(\alpha\) 0.354 vs 0.359 and 0.137 vs 0.133 \\
\end{longtable}
}

In the key numbers, ``a vs b'' is measured against predicted. The D7b row gives the time its predictions were written; its rules were registered with D7. Earlier registrations on the toy tasks (a gate test, PR2 and PR3) concern chunking, passive funnels and a juggling task. The peg gains in Appendix~\ref{app:toy} come from PR3's runs, compared with a window chosen afterwards.

Post hoc analyses, labelled as such in the record and never used for a verdict: the MLP calibration check (Section~\ref{sec:eiv}), the per-axis comparison on Square (Section~\ref{limitations-and-open-questions}), the population-prior bias formula on the MLP and, for both learners, on Square, with the MLP's bootstrap interval there (Section~\ref{sec:bias}), per-state pooling on the MLP, the single-\(V\) comparison and the over-correcting bound (Appendix~\ref{app:second}), and the real-displacement check on the MLP and the success-versus-\(\alpha\) relation in D9 (Section~\ref{sec:closed}). The last was then tested prospectively by D10. The bootstrap intervals for $\rho$ in Figure~\ref{fig:rho} and the Square bias predictions in Table~\ref{tab:bias} were added while writing this paper.

\hypertarget{toy-tasks}{%
\section{Toy tasks, where classical EIV holds}\label{app:toy}}

On two toy tasks the learner's only clean input is the task phase, so the robot state carries no information about the object beyond the phase. There $V$ is the phase-conditional spread of the true error in the demonstrations, and classical EIV and the anchor law coincide. The classical prediction held up on both tasks.

A planar peg insertion and a planar 3-link arm, both in MuJoCo, use scripted experts that correct a noisy reading of the relative error. Per-step MLP behavior cloning sees the noisy frames and the phase. On the arm, the predicted local gain matched the learned gain within 0.13 at all four injection levels, predicted before training. On the peg, where the window that defines ``local'' was chosen, the in-sample match was within 0.09.

A zero-parameter success predictor follows from the gain: simulate the expert with its gain attenuated as predicted. Its first registered test failed (mean absolute error 0.29 over 20 held-out configurations). With the learner protocol then frozen and the arm's depth channel added, it passed at 14 new configurations: with four seeds, mean absolute error 0.10 (95\% CI 0.065 to 0.161) against 0.31 for the expert with the same sensing, paired permutation $p = 0.0008$. It also predicted how history length and noise structure change success (12 configurations, mean absolute error 0.08, Spearman 0.93). Under human-like demonstrations with varied speed, it passed but missed its adoption bar, and a joint-posterior extension failed (Appendix~\ref{the-pre-registration-record}, PR4 to PR9).

\hypertarget{second-order-deviations}{%
\section{Second-order deviations}\label{app:second}}

The small residuals left by the anchor law have the sign that pooling across states predicts (Section~\ref{second-order-pooling-across-states}). For the MLP, resolving the anchor state by state removes most of them. For diffusion, the registered test of that explanation failed narrowly, and a remainder on Square stays open.

\textbf{A shared sign pattern.} In all eight residuals at twice and half the calibration level, the measured ratio lies above the prediction at 24\,mm and below it at 6\,mm. The two residuals at 9 and 18\,mm, though tiny, follow the same pattern (above at 18\,mm, below at 9\,mm; post hoc). This is the pattern Section~\ref{second-order-pooling-across-states} derives for a heterogeneous anchor pooled within each (axis, time) cell. Consistent with it, going from a single $V$ for all cells to the registered one $V$ per cell moves the prediction toward the measurement in all eight cases (post hoc).

\textbf{MLP (post hoc).} Predicting state by state, with each probe state's own $V$, cuts the MLP's residuals on Can from 0.039 and 0.035 to 0.011 and 0.015, and on Square from 0.021 and 0.008 to 0.009 and 0.001 (24 and 6\,mm). The per-state weights reproduce across seeds (median correlation 0.69 on Can, 0.79 on Square). For the MLP, the per-state anchor looks fixed across noise levels, which is the law's assumption at full resolution.

\textbf{Diffusion (registered).} The same explanation was registered for diffusion before its per-state data were analysed, and it missed narrowly on both tasks. On Square, per-state prediction cut the residual by 48\% where the rule required 50\%, and the 6-mm miss fell from 0.094 to 0.050. On Can the registered prediction was already within seed scatter, and the per-state version overshot at 6\,mm (0.510, against 0.527 measured and 0.542 registered).

On Square at 6\,mm, a post hoc bound built to over-correct still leaves a remainder of 0.022 and 0.039 in the two seeds. So the diffusion policy's effective anchor on Square is sharper when it is trained at 6\,mm than at 12\,mm. We report this as an open second-order deviation from the law's assumption that $V$ does not depend on the noise level. A second deviation comes from the benchmark itself. With one frame, the diffusion policies discounted AR(1) noise less than iid noise of the same size (0.273 against 0.187 over all axes), where a fixed $V$ predicts no difference. The test of noise structure within demonstrations found no such effect for the MLP but was not run on diffusion. The first-order law, the pooling correction and these two deviations are separate claims, and only the first was registered as the law.

\end{document}